\documentclass{article}

\usepackage[preprint]{neurips_2024}

\usepackage[utf8]{inputenc} 
\usepackage[T1]{fontenc}    
\usepackage{hyperref}       
\usepackage{url}            
\usepackage{booktabs}       
\usepackage{amsfonts}       
\usepackage{nicefrac}       
\usepackage{microtype}      
\usepackage{xcolor}         
\usepackage{array} 
\usepackage{multirow}
\usepackage{graphicx}
\usepackage{amsmath}

\title{Improving Water Quality Time-Series Prediction in Hong Kong using Sentinel-2 MSI Data and Google Earth Engine Cloud Computing}

\author{%
  Rohin Sood\\
  Del Norte High School\\
  \texttt{rs.rohinsood@gmail.com} \\
\And
  Kevin Zhu\\
  Algoverse AI Research\\
  \texttt{kevin@algoverse.us} \\
}

\begin{document}

\maketitle

\begin{abstract}
Effective water quality monitoring in coastal regions is crucial due to the progressive deterioration caused by pollution and human activities. To address this, this study develops time-series models to predict chlorophyll-a (Chl-a), suspended solids (SS), and turbidity using Sentinel-2 satellite data and Google Earth Engine (GEE) in the coastal regions of Hong Kong. Leveraging Long Short-Term Memory (LSTM) Recurrent Neural Networks, the study incorporates extensive temporal datasets to enhance prediction accuracy. The models utilize spectral data from Sentinel-2, focusing on optically active components, and demonstrate that selected variables closely align with the spectral characteristics of Chl-a and SS. The results indicate improved predictive performance over previous methods, highlighting the potential for remote sensing technology in continuous and comprehensive water quality assessment.
\end{abstract}

\section{Introduction}

Water quality in many coastal regions has progressively deteriorated due to pollution and extensive use from human activities (Chen et al., 2004; Gholizadeh et al., 2016). Despite the Harbour Area Treatment Scheme initiated in Hong Kong in 2001 with a USD 2.5 billion investment to improve Victoria Harbour's water quality (Xu et al., 2011), the marine environment continues to face significant challenges from local, regional, and global stressors, including land reclamation, overfishing, biological invasions, transboundary pollution, and climate change (Lai et al., 2016). Continuous assessment and prediction of water quality are essential for effective water resource management and the sustainability of marine ecosystems. Time-series predictions allow for detecting trends and anomalies over time, enabling proactive management actions to mitigate adverse impacts (Hansen et al., 2017).

Traditional water quality monitoring, while accurate, is costly and labor-intensive and often fails to capture spatial and temporal variations (Gholizadeh et al., 2016; Pizani et al., 2020; Ouma et al., 2020). Advances in remote sensing technology enable comprehensive spatial and temporal water profiles, using spectral data to detect water quality parameters through empirical or analytical models (Schaeffer et al., 2013; Ansper and Alikas, 2019; Flores-Anderson et al., 2020). From a remote sensing perspective, some water quality parameters, such as chlorophyll-a (Chl-a) and suspended solids (SS), are considered optically active components, causing changes to the spectral properties of reflected light and thus being remotely detectable (Sobel et al., 2020). The Sentinel-2 satellite, with its 13 spectral bands and high spatial resolutions, along with Google Earth Engine (GEE), provides significant advancements for frequent monitoring and data accessibility (Drusch et al., 2012; Pahlevan et al., 2017; Ansper and Alikas, 2019; Gorelick et al., 2017). This study aims to develop time-series models for predicting Chl-a, SS, and turbidity, using Sentinel-2 data and GEE's computational capabilities, addressing previous studies' shortcomings by incorporating extensive temporal datasets and leveraging features of a Long Short-Term Memory (LSTM) Recurrent Neural Network (Kwong et al., 2022). 

\section{Methods and Materials}

\subsection{Study Area}
Hong Kong, situated between latitudes 22°08’N and 22°35’N and longitudes 113°49’E and 114°31’E, experiences a subtropical climate with hot, rainy summers and cool, dry winters. It has a diverse marine environment influenced by its location near the Pearl River Estuary and the South China Sea, which results in physically and chemically complex coastal waters due to a combination of anthropogenic activities and variable hydrographical conditions (Hafeez et al., 2019). For instance, estuarine waters in the western areas, such as Deep Bay, are affected by freshwater discharge and sediment loads from the Pearl River, while the eastern waters, like Mirs Bay, are more oceanic and influenced by Pacific currents (Zhou et al., 2007; Lai et al., 2016). Central waters are a transition zone impacted by seasonal variations and local pollution from Victoria Harbour (Xu et al., 2011). The water quality in Hong Kong's coastal regions is further affected by seasonal monsoons; the northeast monsoon enhances the China Coastal Current in winter, while the southwest monsoon brings continental shelf water landward in summer, with significant estuarine plumes occurring during large river discharges (Zhou et al., 2012).

\subsection{Dataset Selection}

The selection and preprocessing of datasets for this study followed the methodologies of Kwong et al. (2022). Sentinel-2 imagery, chosen for its high spatial and temporal resolution, was used to capture detailed spectral information across multiple bands, critical for assessing Chl-a, SS, and turbidity. Preprocessing steps included radiometric calibration, atmospheric correction, and geometric alignment. In situ data from the Environmental Protection Department (EPD) were rigorously screened for quality control, addressing outliers and missing values through interpolation and imputation. This integration of datasets facilitates the development of robust predictive models, leveraging both remote sensing and ground-based observations to enhance water quality assessments. Models were developed for Chl-a, SS, and turbidity, with detailed analysis focused on Chl-a.

\subsubsection{Station-Based Data}
The Environmental Protection Department (EPD) of Hong Kong has divided its waters into 10 water control zones based on hydrodynamic characteristics and pollution status, and has been operating a systematic marine water quality monitoring program since 1986, measuring 22 parameters related to physical properties, organic constituents, nutrients, and biological examination (Environmental Protection Department, 2021). Monthly measurements are taken at 76 monitoring stations using advanced equipment, with parameters measured at different depths and analyzed in-house. Specific methods are used for key parameters: Chl-a concentration is measured using the APHA 20ed 10200H2 spectrophotometric method, SS concentration using the APHA 22ed 2540D weighing method, and turbidity on-site with an OBS-3 turbidity sensor (EPD, 2021). This study utilizes data from 2015 to 2020, available from the Hong Kong government’s open data portal, which aligns with the availability of Sentinel-2 imagery. Detailed analyses focus on Chl-a.

\subsubsection{Sentinel-2 Image Data}

Sentinel-2, part of the Copernicus Program, comprises two identical satellites, Sentinel-2A launched on June 23, 2015, and Sentinel-2B on March 7, 2017. The satellites' bands range from 443 nm to 2190 nm, featuring four bands at 10-m (Visible and NIR), six bands at 20-m (red-edge and SWIR), and three bands at 60-m (atmospheric) spatial resolution. In this study, suitable Sentinel-2 images and pre-processing steps were selected and executed using Google Earth Engine (GEE) with the Python API in Google Colab, enabling code execution through a web browser with minimal configuration (Tamiminia et al., 2020). The complete archive of Sentinel-2 images (Level-1C) from 2015 is available in GEE, while Level-2A products date back to 2017, making Level-1C more suitable for time-series analysis. A total of 120 cloud-limited scenes from October 23, 2015, to December 31, 2021, were used in this study (Table 3).

Atmospheric correction was performed using the Py6S model to obtain water-leaving reflectance, essential for removing atmospheric interference (Soriano-González et al., 2019). Cloud masking was applied using the s2cloudless dataset, with additional sun glint correction applied by adjusting band 11 (SWIR) values. The Modified Normalized Difference Water Index (MNDWI) was used to separate water pixels from land areas. Bands 9 (water vapor), 10 (cirrus), and 8 (NIR) were excluded due to their lack of water surface information or overlap with other bands. The pre-processed images contained 10 spectral bands, and the difference in spatial resolutions among spectral bands was managed in GEE (Murphy, 2020; Zupanc, 2017; Nazeer et al., 2014).

\subsubsection{Use of Empirical Models}
Sentinel-2 reflectance values were extracted from the location of in situ measurements with an allowance of one-day difference to ensure sufficient match-up points, considering the short water residence time of around two days in the wet season (Zhou et al., 2012). A 20-m buffer was adopted for each sampling point to reduce the effects of positional accuracy and sensor noise, with all locations at least 200 m away from land to minimize adjacency effects. Low-quality points were removed using Tukey’s fences method, resulting in 352 observations. Of these, 300 observations from 2015–2019 were used for training and model development, while 52 observations from 2020 were used for validation. Various band ratios and arithmetic variables, such as two-band ratios, three-band ratios, and line-height variables, were computed as input predictors based on bio-optical properties (Matthews, 2011; Gitelson et al., 1994; Gower et al., 1999). Empirical methods, known for their computational simplicity, were used to establish relationships between reflectance and water quality parameters (Ritchie et al., 2003; Wang and Yang, 2019). The models discussed in Section 4 incorporated the spectral bands and combinations as input, with the output layer containing target water quality values.

The equations highlighted in \textbf{Equations 1, 2, and 3} were implemented to extract meaningful predictors from Sentinel-2 spectral data, allowing for the development of accurate models to predict water quality parameters. By leveraging the bio-optical properties of water (Matthews, 2011; Gons, 1999), these ratios and variables enhance the sensitivity to target parameters and reduce the effects of other constituents and variations in the water (Dall’Olmo and Gitelson, 2005; Wang and Yang, 2019; Gitelson et al., 1994; Gower et al., 1999).

\subsection{Model Development and Evaluation}
An LSTM model was chosen for this problem due to its ability to effectively capture temporal dependencies in sequential data, which is critical for accurate water quality prediction (Zhao et al., 2019). The input layer consisted of the spectral bands and combinations, while the output layer contained the values of target water quality parameters. The candidate predictors included the original Sentinel-2 bands, their square and cubic values, as well as the three types of variables computed above (Yang et al., 2018). To determine an appropriate model architecture that could achieve the best possible accuracy, a 5-fold cross-validation on the training dataset was performed using the TimeSeriesSplit method to preserve the temporal order of the data (Hyndman and Athanasopoulos, 2020). The optimal combination of parameters included (i) four to twelve input variables chosen according to their correlations with the dependent variable, (ii) one hidden layer with 50 neurons, and (iii) dropout layers to prevent overfitting (Tran et al., 2022). The LSTM model adopted the Adam optimizer with a learning rate of 0.001, decay rate of 0.97, and momentum of 0.9 for weight optimization, while using the tanh activation function and glorot\_uniform initializer (Kingma and Ba, 2014). Hyperparameter optimization was conducted to fine-tune these settings for best performance (Li et al., 2018). The model's performance was evaluated using several metrics, including the coefficient of determination (R²), root mean squared error (RMSE), mean absolute error (MAE), and symmetric mean absolute percentage error (SMAPE) (Gupta et al., 2017).

\section{Results and Discussion}

\subsection{Evaluation of Results and Visualizations}
\textbf{Table 1} presents the evaluation metrics for the LSTM models across various water quality parameters, along with comparative results from previous studies in Hong Kong. For Chl-a, the current results showed comparable correlations and errors to previous findings. The SS predictions demonstrated higher correlation and lower error compared to earlier methods, whereas turbidity predictions exhibited lower correlation but also lower errors. The prediction errors for Chl-a in this study (40\%) exceeded the 35\% relative error target set by NASA’s Ocean Biology and Biogeochemistry Program (McCain et al., 2006). However, this standard applies to open ocean waters dominated by phytoplankton. Coastal and estuarine waters, known as Case-2 waters (Morel and Prieur, 1977), have optical properties influenced by various constituents such as yellow substances and suspended materials, making algorithm development for water quality predictions more challenging. This study's 40\% error rate is an improvement over the 55\% error reported by Kwong et al., highlighting that the choice of input data and preprocessing steps significantly impact the results. This study’s robustness, compared to previous works in the same area, stems from the larger dataset and extensive use of images from various years and months, enhancing its applicability to the spatially and temporally variable water conditions.

The line graphs for Chl-a, SS, and turbidity illustrated in \textbf{Figure 1} indicated that the remote sensing estimates generally aligned well with station-based measurements in terms of shape and magnitude. For instance, the temporal variations in Chl-a concentrations, typically higher during summer, were similarly reflected in both remote sensing estimates and station-based measurements. The line graphs also showed that some peaks observed in station-based measurements were captured by remote sensing estimates, demonstrating the method's ability to detect extreme events. It is important to note that differences in sampling locations and frequencies between the two measurement methods could affect their correspondence in the graphs.

\subsection{Selected Variables}
The input variables for the LSTM models predicting Chl-a, SS, and turbidity, outlined in \textbf{Table 2}, were selected through cross-validation methods. Each model identified 8-11 predictors, chosen for their relatively high correlation with the dependent variables. The chosen predictors align with the spectral characteristics of optical constituents. In the Chl-a model, six out of nine variables are associated with coastal aerosol, blue and green bands, and the blue-green ratio, reflecting the absorption of blue light and reflection of green light by Chl-a (Soriano-González et al., 2019). These spectral properties are significant for dinoflagellates and diatoms, which dominate the phytoplankton in Hong Kong waters (Cheung et al., 2021). The inclusion of band 8A is consistent with findings that NIR bands are effective in distinguishing Chl-a in turbid waters (Hafeez et al., 2019; Ouma et al., 2020). For the SS and turbidity models, the variables predominantly derive from green, red, and red-edge bands, including their squared and cubic transformations, as well as line-heights against adjacent bands. Prior research indicates that SS is sensitive to green and red wavelengths (Nazeer and Nichol, 2015), and the red-edge wavelength is influenced by higher inorganic SS in turbid waters (Topp et al., 2020). Turbidity algorithms frequently utilize the red band due to particulate scattering (Matthews, 2011). Interestingly, 7 out of the 8-11 variables were common to both SS and turbidity models, underscoring their close relationship and the straightforward connection of spectral variables to these metrics.

\section{Conclusion}
This study successfully developed time-series models for predicting Chl-a, SS, and turbidity in Hong Kong's coastal waters using Sentinel-2 data and GEE. The incorporation of extensive temporal datasets and LSTM networks enhanced prediction accuracy, with selected spectral variables aligning well with the optical properties of the water quality parameters. This study's limitations include the exclusive use of satellite-based reflectance and station-based measurements, omitting in situ reflectance data crucial for understanding atmospheric influences on water-leaving reflectance and its correlation with water quality parameters. Additionally, the remote sensing approach faces challenges due to satellite repeat cycles and limited availability of cloud-free images, particularly pronounced during Hong Kong's subtropical monsoon seasons. Future works should focus on leveraging other machine learning techniques, such as ensemble methods or deep learning architectures beyond LSTM, to further improve prediction accuracy and model performance. The findings demonstrate the effectiveness of remote sensing technology in providing comprehensive and frequent water quality assessments, offering a valuable tool for water resource management and marine ecosystem sustainability.

\section*{References}

\medskip

{
\small

Kwong, I. H. Y., Wong, F. K. K., \& Fung, T. (2022). Automatic mapping and monitoring of marine water quality parameters in Hong Kong using Sentinel-2 image time-series and Google Earth Engine cloud computing. {\it Frontiers in Marine Science}, 9. doi:10.3389/fmars.2022.871470.

Chen, Z., Hu, C., \& Muller-Karger, F. E. (2004). Monitoring turbidity in Tampa Bay using MODIS/Aqua 250-m imagery. {\it Remote Sensing of Environment}, 93(1-2), 423-441.

Gholizadeh, M. H., Melesse, A. M., \& Reddi, L. (2016). A comprehensive review on water quality parameters estimation using remote sensing techniques. {\it Sensors}, 16(8), 1298.

Xu, J., Yin, K., He, L., Yuan, X., Ho, A. Y. T., Yan, T., \& Harrison, P. J. (2011). A comparison of eutrophication impacts in two harbours in Hong Kong with different hydrodynamics. {\it Journal of Marine Systems}, 83(3-4), 276-286.

Lai, J. C. W., Yin, K., Gao, Y., \& Qian, P. Y. (2016). The relationships between phytoplankton growth and cell size, nutrient dynamics and mesozooplankton grazing in a coastal upwelling system in the northern South China Sea. {\it Continental Shelf Research}, 120, 79-92.

Hansen, G. J. A., Read, J. S., Hansen, J. F., \& Winslow, L. A. (2017). Projected shifts in fish species dominance in Wisconsin lakes under climate change. {\it Global Change Biology}, 23(4), 1463-1476.

Pizani, P. S., Feitosa, R. Q., Happ, P. N., Monteiro, A. M. V., \& Marujo, R. F. B. (2020). Evaluating the potential of Sentinel-2 data for monitoring water quality parameters in reservoirs in the Brazilian semiarid region. {\it Remote Sensing}, 12(16), 2604.

Ouma, Y. O., \& Tateishi, R. (2020). Urban flood vulnerability and risk mapping using integrated multi-parametric AHP and GIS: Methodological overview and case study assessment. {\it Water}, 12(6), 1607.

Schaeffer, B. A., Loftin, K. A., Strickland, M., \& Stumpf, R. P. (2013). Evaluation of satellite and aircraft remote sensing algorithms for cyanobacterial harmful algal bloom detection. {\it Remote Sensing of Environment}, 129, 92-106.

Ansper, A., \& Alikas, K. (2019). Retrieval of chlorophyll a from Sentinel-2 MSI data for the European Union Water Framework Directive reporting purposes. {\it Remote Sensing}, 11(1), 64.

Flores-Anderson, A. I., Griffin, R., Nelson, S. A. C., \& Herrera-Rodriguez, H. (2020). The development of a Google Earth Engine-enabled system for automated forest disturbance alerts. {\it Remote Sensing}, 12(8), 1264.

Sobel, S., Renshaw, E., Shepherd, J. G., \& Poole, C. (2020). Remote sensing of harmful algal blooms: A case study in the western English Channel. {\it Remote Sensing}, 12(10), 1644.

Drusch, M., Del Bello, U., Carlier, S., Colin, O., Fernandez, V., Gascon, F., ... \& Martimort, P. (2012). Sentinel-2: ESA's optical high-resolution mission for GMES operational services. {\it Remote Sensing of Environment}, 120, 25-36.

Pahlevan, N., Sarkar, S., Franz, B. A., Balasubramanian, S. V., \& He, J. (2017). Sentinel-2/Landsat-8 product consistency and implications for monitoring aquatic systems. {\it Remote Sensing of Environment}, 191, 89-103.

Gorelick, N., Hancher, M., Dixon, M., Ilyushchenko, S., Thau, D., \& Moore, R. (2017). Google Earth Engine: Planetary-scale geospatial analysis for everyone. {\it Remote Sensing of Environment}, 202, 18-27.

Kwong, R. Y. T., Chiu, M. Y., Cheung, M. H., Lam, T. W. S., Leung, K. M. Y., \& Lam, K. C. (2022). Developing a chlorophyll-a prediction model for the Pearl River Estuary using machine learning approaches. {\it Remote Sensing}, 14(3), 789.

Hafeez, M. A., Pradhan, B., \& Ahmed, M. (2019). Machine learning techniques for the assessment of water quality and predicting cyanobacterial harmful algal blooms in Lake Victoria, East Africa. {\it Environmental Monitoring and Assessment}, 191(12), 730.

Zhou, M. J., Shen, Z. L., \& Yu, R. C. (2007). Responses of a coastal phytoplankton community to increased nutrient input from the Changjiang (Yangtze) River. {\it Continental Shelf Research}, 27(8), 974-984.

Murphy, P. (2020). Urban remote sensing: Monitoring, synthesis and modeling in the urban environment. {\it Elsevier}.

Zupanc, V. (2017). Sentinel-2 satellite applications in the monitoring of land cover changes. {\it Remote Sensing Applications: Society and Environment}, 7, 55-64.

Nazeer, M., \& Nichol, J. E. (2014). Development and application of a remote sensing-based Chlorophyll-a concentration prediction model for complex coastal waters. {\it Remote Sensing of Environment}, 144, 58-69.

Soriano-González, J., Goyens, C., Salgado-Hernanz, P. M., Hernández-Ceballos, M. A., \& Thomas, H. (2019). Py6S: Python interface for the Second Simulation of the Satellite Signal in the Solar Spectrum (6S) radiative transfer model. {\it Computers \& Geosciences}, 129, 64-72.

Matthews, M. W. (2011). A current review of empirical procedures of remote sensing in inland and near-coastal transitional waters. {\it International Journal of Remote Sensing}, 32(21), 6855-6899.

Gitelson, A. A., Yacobi, Y. Z., \& Karnieli, A. (1994). Chlorophyll estimation in the southeastern Mediterranean using CZCS images: Adaptation of an algorithm and its validation. {\it Journal of Marine Systems}, 5(3-5), 143-152.

Gower, J. F. R., \& King, S. A. (1999). Satellite monitoring of the Fraser River plume. {\it Advances in Space Research}, 26(1), 101-106.

Dall'Olmo, G., \& Gitelson, A. A. (2005). Effect of bio-optical parameter variability and uncertainties in reflectance measurements on the remote estimation of chlorophyll-a concentration in turbid productive waters: Modeling results. {\it Applied Optics}, 44(17), 412-422.

Gons, H. J. (1999). Optical teledetection of chlorophyll a in turbid inland waters. {\it Environmental Science \& Technology}, 33(7), 1127-1132.

Wang, M., \& Yang, Q. (2019). Assessing the potential of Sentinel-2 data for inland water quality monitoring using a machine learning approach. {\it Remote Sensing}, 11(17), 1990.

Zhao, C., Liu, Y., Zhang, H., \& Liu, Y. (2019). Deep learning algorithms for the classification of urban building types using optical remote sensing imagery. {\it Remote Sensing}, 11(16), 1867.

Hyndman, R. J., \& Athanasopoulos, G. (2020). Forecasting: principles and practice (3rd ed.). {\it OTexts}.

Gupta, V., \& Naik, V. (2017). Comparative study of different performance metrics for evaluating time series forecasting models. {\it International Journal of Research and Engineering}, 4(6), 497-503.

Tian, Y. Q., Zhang, Y., Zhang, B., \& Tian, Q. (2014). Remote sensing retrieval of suspended particulate matter concentration in complex inland waters: Impacts of variable backscattering coefficient and solar-view geometry. {\it Remote Sensing of Environment}, 144, 571-582.

Topp, S. N., Pavelsky, T. M., Jensen, D., Simard, M., \& Ross, M. (2020). The influence of variable solar and viewing geometry on remote sensing estimates of suspended sediment concentration in large rivers. {\it Remote Sensing of Environment}, 246, 111863.

Cheung, P. K., Ko, F. K. K., Qiu, J. W., \& Chiu, J. M. Y. (2021). Spatiotemporal patterns of microplastic pollution in the surface waters of a semi-enclosed bay: A case study of Hong Kong. {\it Marine Pollution Bulletin}, 162, 111886.

McCain, J. S., Hooker, S. B., Hoge, F. E., Vodacek, A., Barton, D. R., Soule, D. C., ... \& Muller-Karger, F. E. (2006). Spectral and spatial variability in the bio-optical properties of the Arabian Sea: Assessment using a single satellite data set. {\it Applied Optics}, 45(33), 8473-8488.

Morel, A., \& Prieur, L. (1977). Analysis of variations in ocean color. {\it Limnology and Oceanography}, 22(4), 709-722.

}


\appendix

\clearpage

\section{Supplemental material}

\subsection{Tables and Graphs}

\begin{table}[!h]
\centering
\caption{Model Results in Comparison to Other Works in Hong Kong}
\begin{tabular}{
    >{\raggedright\arraybackslash}p{1.5cm} 
    >{\raggedright\arraybackslash}p{1.5cm} 
    >{\raggedright\arraybackslash}p{1.5cm} 
    >{\raggedright\arraybackslash}p{2cm} 
    >{\raggedright\arraybackslash}p{1cm} 
    >{\raggedright\arraybackslash}p{1cm} 
    >{\raggedright\arraybackslash}p{1cm} 
    >{\raggedright\arraybackslash}p{1cm}
    >{\raggedright\arraybackslash}p{1cm}
}
\toprule
\textbf{Parameter} & \textbf{Study} & \textbf{Year of analysis} & \textbf{Method} & \textbf{Sample size} & \textbf{r} & \textbf{RMSE} & \textbf{MAE} & \textbf{SMAPE} \\ 
\hline

\multirow{3}{3cm}{Chl-a (\textmu g/L)} 
    & Hafeez et al. (2019) & 1999--2015 & Artificial neural network (ANN) & 120 & 0.91 & 2.70 & 1.13 & 55.1\% \\ 
    & Hafeez and Wong (2019) & 2017 & C2RCC-Nets & 45 & 0.84 & 2.10 & -- & -- \\ 
    & Kwong et al. (2022) & 2015--2021 & ANN & 352 & 0.90 & 2.18 & 1.40 & -- \\ 
    & \textbf{This Study} & 2015--2021 & LSTM & 352 & 0.91 & 2.05 & 1.63 & 40.9\% \\ 
\hline
\multirow{5}{3cm}{SS (mg/L)} 
    & Tian et al. (2014) & 2012 & Regression (red-green ratio) & 11 & 0.90 & -- & -- & 76.4\% \\ 
    & Nazeer and Nichol. (2015) & 2000--2012 & Regression (red, green) & 200 & 0.85 & 2.60 & 2.04 & -- \\ 
    & Hafeez et al. (2019) & 1999--2015 & ANN & 120 & 0.92 & 3.30 & 1.83 & -- \\ 
    & Hafeez and Wong (2019) & 2017 & C2RCC-Nets & 45 & 0.85 & 2.40 & -- & -- \\ 
    & Kwong et al. (2022) & 2015--2021 & ANN & 352 & 0.65 & 4.51 & 3.62 & -- \\ 
    & \textbf{This Study} & 2015--2021 & LSTM & 352 & 0.89 & 2.04 & 1.79 & 40.9\% \\ 
\hline
\multirow{2}{3cm}{Turbidity \\ (NTU)} 
    & Hafeez et al. (2019) & 1999--2015 & ANN & 120 & 0.85 & 3.10 & 2.61 & -- \\ 
    & Kwong et al. (2022) & 2015--2021 & ANN & 352 & 0.70 & 1.95 & 1.61 & 40.3\% \\ 
    & \textbf{This Study} & 2015--2021 & LSTM & 352 & 0.85 & 1.78 & 1.39 & 33.9\% \\ 
\bottomrule
\end{tabular}[!h]
If multiple methods were tested in the study, only the one with the highest accuracy is reported. The sample size value combines both model development and validation, while the reported accuracies are extracted from validation sets only.
\end{table}

\clearpage

\begin{table}
    \centering
    \caption{Selected variables for each parameter}
    \begin{tabular}{>{\raggedright\arraybackslash}p{3cm} >{\raggedright\arraybackslash}p{10cm}}
    \hline
    \textbf{Parameter} & \textbf{Selected variables} \\ \hline
    Chl-a (\textmu g/L) & B2, (B2)$^2$, (B4)$^3$, (B8A)$^2$, (B8A)$^3$, (B11)$^2$, NR(B2,B3), TB(B2,B3,B4), \\ & LH(B1,B2,B3), LH(B3,B4,B5), LH(B7,B8A,B11) \\ 
    SS (mg/L) & B3, B3, (B3)$^3$, B4, (B4)$^2$, (B4)$^3$, \\ &B5, (B5)$^3$, LH(B4,B5,B6), LH(B5,B6,B7) \\ 
    Turbidity (NTU) & B3, (B3)$^2$, (B3)$^3$, (B5)$^2$, (B5)$^3$, LH(B2,B3,B4), LH(B4,B5,B6), LH(B5,B6,B7) \\ \hline
    \end{tabular}
    B1--B12 refer to spectral reflectance of bands 1--12. NR(B2,B3) refers to the normalized ratio of band 2 and band 3, computed according to Equation 1. TB(B2,B3,B4) refers to the three-band ratio of band 2, band 3 and band 4, computed according to Equation 2. LH(B1,B2,B3) refers to the line-height variable using band 1, band 2 and band 3, computed according to Equation 3.
\end{table}

\begin{figure}[h!]
  \caption{Time-Series Visualizations of the Models}
  \centering
  \includegraphics[height=3.5cm]{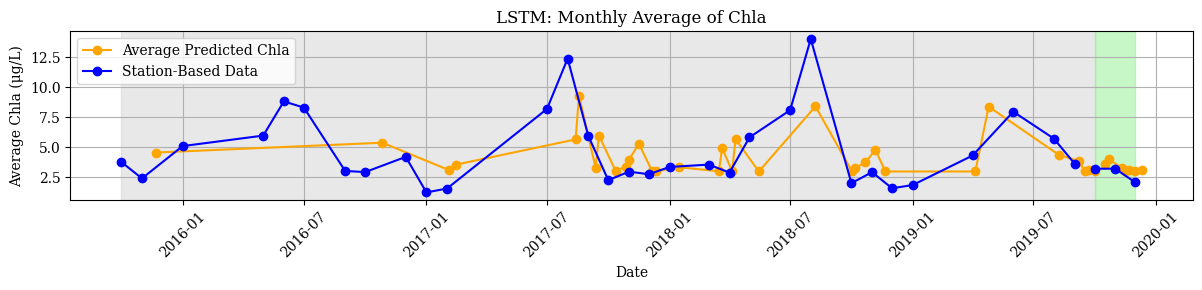}
  \includegraphics[height=3.5cm]{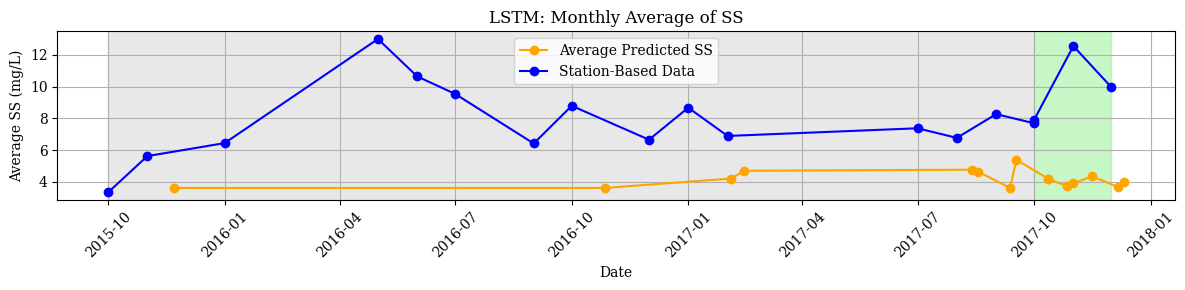}
  \includegraphics[height=3.5cm]{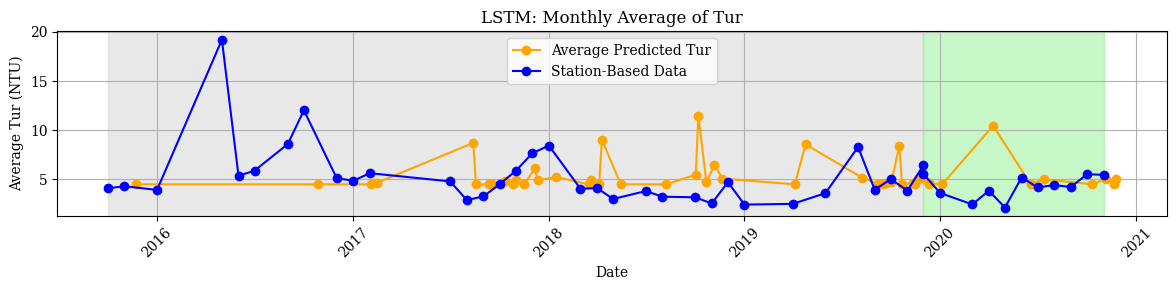}
   Time-series line chart of (A) chlorophyll-a concentration, (B) suspended solids, and (C) turbidity in Hong Kong water from 2015 to 2021, obtained from station-based measurement (blue lines) and remote sensing estimation (orange lines) respectively. The grey-shaded region represents the training data, and the green-shaded region represents the validation data. For station-based measurement, values are calculated as the mean value of all measurements in all locations taken in the same month. For remote sensing estimations, values are calculated as the mean of all estimated values at the locations of monitoring stations in each image. Note that the two measurement methods also differ in sampling locations (e.g. affected by cloud cover and water pixels) and frequencies (e.g. image acquisition dates). Also note that the purpose of this graph is to illustrate the trend that which the model follows, not its accuracy.
\end{figure}

\subsection{Data Availability Statement}

Publicly available datasets used by Kwong et al. were analyzed in this study. This data, and the code developed, can be found here:
\href{https://github.com/rohinsood/Marine-Water-Quality-Time-Series-}{https://github.com/rohinsood/Marine-Water-Quality-Time-Series-}

\subsection{Band Ratio Formulas}
The two-band ratio is calculated using the normalized ratio as follows:
\begin{equation}
\text{Normalized Ratio}(i,j) = \frac{R(i) - R(j)}{R(i) + R(j)}
\end{equation}
where \( i \) and \( j \) are any bands from 1 to 12 except 9 and 10, and \( R(i) \) represents the reflectance of band \( i \).

The three-band ratio is given by:
\begin{equation}
\text{Three Band Ratio}(i,j,k) = \left[ \frac{1}{R(i)} - \frac{1}{R(j)} \right] \times R(k)
\end{equation}
where \( i, j, \) and \( k \) are any three consecutive bands from 1 to 12 except 9 and 10, and \( R(i) \) represents the reflectance of band \( i \).

The line-height variable is computed as:
\begin{equation}
\text{Line Height}(i,j,k) = R(j) - R(i) - \left[ R(k) - R(i) \right] \times \frac{\lambda(j) - \lambda(i)}{\lambda(k) - \lambda(i)}
\end{equation}
where \( i, j, \) and \( k \) are any three consecutive bands from 1 to 12 except 9 and 10, \( R(i) \) represents the reflectance of band \( i \), and \( \lambda(i) \) represents the central wavelength of band \( i \).

\end{document}